\documentclass[letterpaper, 10 pt, conference]{ieeeconf}  

\IEEEoverridecommandlockouts                              

\usepackage{cite}
\usepackage{amsmath,amssymb,amsfonts}
\usepackage{algorithmic}
\usepackage{graphicx}
\usepackage{textcomp}
\usepackage{xcolor}

\usepackage{algorithm}

\usepackage{color}
\newcommand{\nop}[1]{}
\usepackage{booktabs}
\usepackage{multirow}
\usepackage{makecell}
\usepackage{amsmath}
\usepackage{subfigure}

\title{\LARGE \bf
TOFG: A Unified and Fine-Grained Environment Representation in Autonomous Driving
}

\author{Zihao Wen$^{1*}$, Yifan Zhang$^{1*}$, Xinhong Chen$^{1}$, and Jianping Wang$^{1}$
\thanks{*The authors contributed equally to this work. }
\thanks{$^{1}$Zihao Wen, Yifan Zhang, Xinhong Chen, and Jianping Wang are with Department of Computer Science,
        City University of Hong Kong, Hong Kong, China, and City University of Hong Kong Shenzhen Research Institute, Shenzhen, China
        (Email: zihaowen2-c@my.cityu.edu.hk, \{yif.zhang, xinhong.chen, jianwang\}@cityu.edu.hk).}%
\thanks{This work was partially supported by Hong Kong Research Grant Council under GRF 11200220, Science and Technology Innovation Committee Foundation of Shenzhen under Grant No. JCYJ20200109143223052. }
}

\begin{document}

\maketitle
\thispagestyle{empty}
\pagestyle{empty}

\begin{abstract}
In autonomous driving, an accurate understanding of environment, e.g., the vehicle-to-vehicle and vehicle-to-lane interactions, plays a critical role in many driving tasks such as trajectory prediction and motion planning. Environment information comes from high-definition (HD) map and historical trajectories of vehicles. Due to the heterogeneity of the map data and trajectory data, many data-driven models for trajectory prediction and motion planning extract vehicle-to-vehicle and vehicle-to-lane interactions in a separate and sequential manner. However, such a manner may capture biased interpretation of interactions, causing lower prediction and planning accuracy. Moreover, separate extraction leads to a complicated model structure and hence the overall efficiency and scalability are sacrificed. To address the above issues, we propose an environment representation, Temporal Occupancy Flow Graph (TOFG). Specifically, the occupancy flow-based representation unifies the map information and vehicle trajectories into a homogeneous data format and enables a consistent prediction. The temporal dependencies among vehicles can help capture the change of occupancy flow timely to further promote model performance. To demonstrate that TOFG is capable of simplifying the model architecture, we incorporate TOFG with a simple graph attention (GAT) based neural network and propose TOFG-GAT, which can be used for both trajectory prediction and motion planning. Experiment results show that TOFG-GAT achieves better or competitive performance than all the SOTA baselines with less training time.



\end{abstract}


\section{Introduction} \label{sec:intro}

Autonomous driving has gained rapid development in recent years. In a typical autonomous driving system, there are three indispensable modules~\cite{yurtsever2020survey,grigorescu2020survey}, i.e., perception, planning, and control modules. The planning module contains two major tasks, namely, trajectory prediction of surrounding vehicles and motion planning of the ego vehicle. Both trajectory prediction and motion planning take the HD map and historical trajectories of \nop{surrounding }vehicles as the environment inputs. Such two types of data are necessary to\nop{ represent and} capture interactions among vehicles and lanes, which largely determine the performance of the aforementioned tasks~\cite{ye2022gsan,interaction1,interaction2}. 


Given the great capability of capturing relations among agents, Graph Neural Network (GNN)~\cite{velickovic2017gat} based models have been widely used for trajectory prediction~\cite{zhou2022hivt,liu2021multimodal,Gu_2021_ICCV,ngiam2021scene} and motion planning tasks~\cite{NEURIPS2021_224e5e49,interaction-planning,liang2020lanegcn}. Most of the GNN-based models adopt attention mechanisms~\cite{vaswani2017attention} to capture the interactions among agents, which helps improve the model performance. There are two main types of interaction that are commonly considered in the literature, namely, vehicle-to-vehicle interactions~\cite{learn-planning1,v2v2,v2v3,gohome} and vehicle-to-lane interactions~\cite{v2l1,Gao_2020_CVPR,Chang_2019_CVPR}. These two types of interactions are extracted from the sequential historical trajectories and the graph structured HD map. In the literature, different attention layers are designed manually to represent different types of interactions~\cite{liang2020lanegcn,interaction3}. The manually designed attention order deems to have some impact on the output of the model since different directions and orders of data flow lead to different results. Such approaches create the following issues. 
\begin{itemize}
    \item Firstly, it is hard to design a one-size-fits-all attention order manually, which can work well in all driving scenarios. 
    \item
    Secondly, designing dedicated attention layers for every type of interactions results in a complex network architecture, which decreases the computation efficiency and sacrifices the scalability of the model. 
    \item
    Thirdly, due to the already complex attention mechanism design, most current GNN models adopt a coarse-grained map structure to simplify the graph for easy training. However, such a manner may lead to the loss of attention in some regions and further cause the decrease of the overall model performance.
\end{itemize}

To address the above issues, both vehicle-to-vehicle and vehicle-to-lane interactions need to be captured with fine-grained map information simultaneously. To this end, a unified and fine-grained representation that expresses the HD map and the trajectory of surrounding vehicles in an isomorphic way is urged to be explored.

The recently proposed Occupancy Flow Field (OFF)~\cite{mahjourian2022occupancy} is a potential solution that has been utilized to provide a unified representation of the environment for predicting traffic from a macroscopic view. OFF, however, neglects the microscopic interaction and trajectory of a single vehicle. In this paper, we adopt and adapt OFF to construct a \textbf{Temporal Occupancy Flow Graph (TOFG)} representation that unifies the HD map information and historical trajectories of vehicles in an \nop{occupancy flow graph}\textcolor{black}{isomorphic data format}.
Specifically, the information of lanes, vehicles, and other \nop{traffic infrastructures}\textcolor{black}{road information} are represented as nodes in the TOFG. Spatial and temporal edges are then created among \nop{vehicles and lanes}\textcolor{black}{them} to simultaneously capture the vehicle-to-vehicle and vehicle-to-lane interactions in a consistent manner for more accurate prediction results.
Such a unified representation can simplify the GNN structure and further help to improve computation efficiency and scalability. Moreover, the simplification of TOFG also leaves room to enable a fine-grained map structure. 
Our main contributions are summarized as follows:
\begin{itemize}
    \item 
    We propose TOFG to completely characterize the heterogeneous environment in a unified way and simultaneously capture vehicle-to-vehicle and vehicle-to-lane interactions. Both the historical trajectories of surrounding vehicles and the HD maps are encoded with fine-grained representations to include the temporal and spatial environment information \textcolor{black}{effectively and }efficiently.
    \item 
    We apply a simple Graph Attention (GAT) based neural network with TOFG, referred to as \textbf{TOFG-GAT}, to showcase that our proposed TOFG is capable of simplifying the model architecture without sacrificing performance. Specifically, compared with several SOTA models, the extensive experiment results show that TOFG-GAT can achieve better performance in both trajectory prediction and motion planning with \textbf{68.57\%} fewer parameters and \textbf{33.69\%} less training time. 
    \item 
    We visualize the attention map of TOFG-GAT and other baselines. The results show that TOFG-GAT produces a more effective and rational driving interaction logic of the model.
\end{itemize}
The rest of the paper is organized as follows. In Section~\ref{sec:TOFG}, we use an example to motivate the necessity of unified representation and then present the architecture of our proposed TOFG for environment representation. We then elaborate on the details of the TOFG-GAT for trajectory prediction and motion planning tasks in Section~\ref{sec:model}. We conduct extensive experiments in Section~\ref{sec:experiment} to demonstrate the superiority of our proposed TOFG in representing environments. Finally, we conclude this paper in Section~\ref{sec:conclusion}.

\section{Temporal Occupancy Flow Graph} \label{sec:TOFG}

In this section, we present our unified environment representation TOFG. To begin with, we use a motivation example to illustrate the limitations of separately extracting vehicle-to-vehicle and vehicle-to-lane interactions. Then, we introduce how a TOFG can be constructed based on the input of HD maps and historical vehicle trajectories in details.

\subsection{Motivation Example} \label{sec:motivation_example}

\begin{figure}[h]
    \centering
    \includegraphics[width=0.95\linewidth]{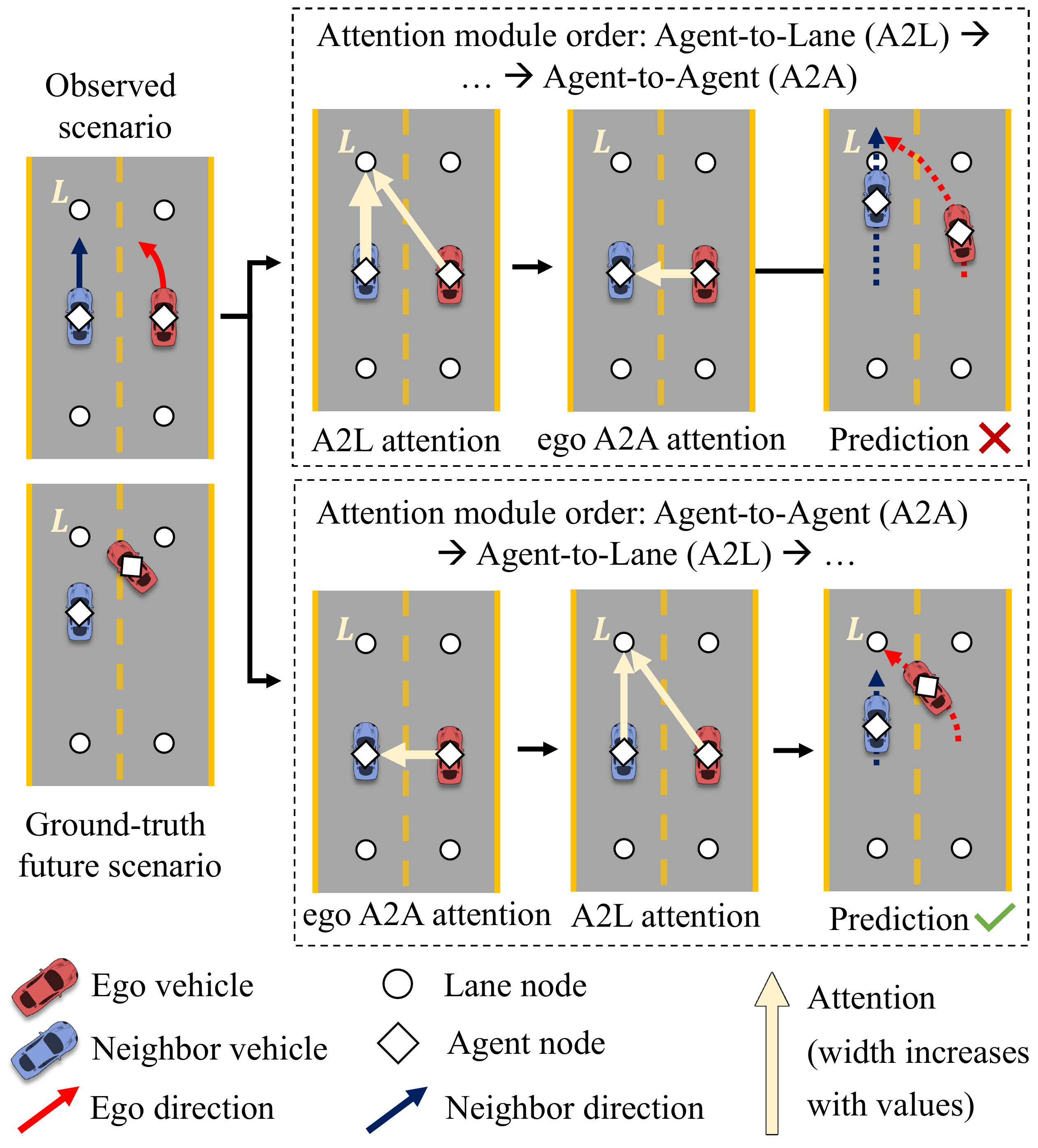}
    \vspace{-0.3cm}
    \caption{A failure case of LaneGCN. Manually designed attention order could not handle vehicle-to-vehicle and vehicle-to-lane interactions comprehensively.}
    \label{fig:lgcn_weak_case}
    \vspace{-0.7cm}
\end{figure}


\nop{Supported by the latest autonomous datasets~\cite{}, many GNN-based models now take agent trajectories and HD maps as input when conducting trajectory prediction task or motion planning task \cite{zhou2022hivt,ye2022gsan,liang2020lanegcn}. Specifically, trajectory data is usually modeled as a sequence of states of the vehicle and HD map is formatted as a graph with lane segments being nodes, resulting in the heterogeneity of these two data. Thereupon, previous models such as LaneGCN~\cite{liang2020lanegcn} and HiVT~\cite{zhou2022hivt} utilize multiple attention modules to capture the interactions between dynamic vehicles and static lanes or the ones among vehicles separately.}

To motivate the necessity of unified representation, we take LaneGCN~\cite{liang2020lanegcn} as an example, which uses three GATs, namely Actor-to-Lane (A2L), Lane-to-Actor (L2A), and Actor-to-Actor (A2A), to capture the interactions among vehicles and lanes. Specifically, A2L aggregates vehicle information to lanes, L2A passes fused traffic information to vehicles, and A2A handles the interaction between vehicles. 
\nop{However, the model's understanding of scenario context is different given different order of attention modules and further leads to inconsistent prediction. }
We use a simple overtaking driving scenario to demonstrate that the order of attention layers does have impacts on capturing the interaction and the prediction accuracy, as shown in Figure~\ref{fig:lgcn_weak_case}. 

In Figure~\ref{fig:lgcn_weak_case}, the red ego vehicle is overtaking its blue neighbor vehicle, and $L$ is a lane node representing the lane segment in front of the neighbor vehicle. Consider two different sequential orders of LaneGCN's attention layers: the first one is \textless A2L, L2A, A2A\textgreater, which is the same as the order in~\cite{liang2020lanegcn}, and the second one is \textless A2A, A2L, L2A\textgreater\nop{to put A2A before A2L}. In the first case, A2L is computed first, so the neighbor vehicle has a higher probability of appearing at $L$ since it is closer. 
\textcolor{black}{Then, L2A passes such information of $L$ to two vehicles.
In the following A2A, the red vehicle may abort overtaking since it has received the information about the blue vehicle from $L$. Thus, in A2A, the red vehicle node sends weak lane overtaking intention to the blue vehicle node, leading to wrong prediction results.} \nop{given the context of understanding A2L, the ego vehicle is likely to abort overtaking at node $L$ since the information of the neighbor vehicle soon occupying $L$ is passed to the ego one.} \nop{it will be occupied by the neighbor vehicle soon.} In the second case, A2A is computed first, and thus only the current information between vehicles is exchanged. When A2L is performed, the red vehicle will have no information that the blue vehicle may occupy node $L$. Hence, the red vehicle may still have a higher probability of appearing at node $L$.
\textcolor{black}{In the following L2A, the blue vehicle receives the information that the red vehicle will occupy $L$, and thus it may give the road.}
 Based on the above two cases, either order of attention modules could not handle the scenario comprehensively since they do not compute vehicle interaction and lane occupation simultaneously. \textcolor{black}{Generally speaking, \nop{the vehicle nodes and lane nodes are updated in attention layers designed to process V2L or V2V interactions. If the update for V2V and the update for V2L happen in a sequential manner, then in the prior update, only one kind of interaction is considered. Therefore, biased interpretation may possibly be incorporated in vehicle nodes and lane nodes, leading to a wrong understanding of the environment, thus, causing a decrease in prediction accuracy.} a fixed attention order brings biased understanding of the environment, leading to a decrease of prediction accuracy.}
\textcolor{black}{To verify the existence of above issues, we conduct experiments to compare the prediction accuracy of the aforementioned two LaneGCN models with different attention layer orders\nop{: the vanilla LaneGCN (attention layer order: \textless A2L, L2A, A2A\textgreater) and the modified LaneGCN with A2A before A2L (attention layer order: \textless A2A, A2L, L2A\textgreater)}. Please refer to Section~\ref{sec:traj_pred} for details.}

To tackle the above issues, we propose Occupancy Flow Graph (OFG) and Temporal Occupancy Flow Graph (TOFG), two vectorized representations combining HD maps and vehicle trajectory information\nop{ Our OFG and TOFG integrate trajectory (sequence data) and HD map (graph data) to create a homogeneous data representation}, which enables capturing more accurate interactions simultaneously.



\begin{figure*}[h]
    \centering
    \includegraphics[width=0.95\textwidth]{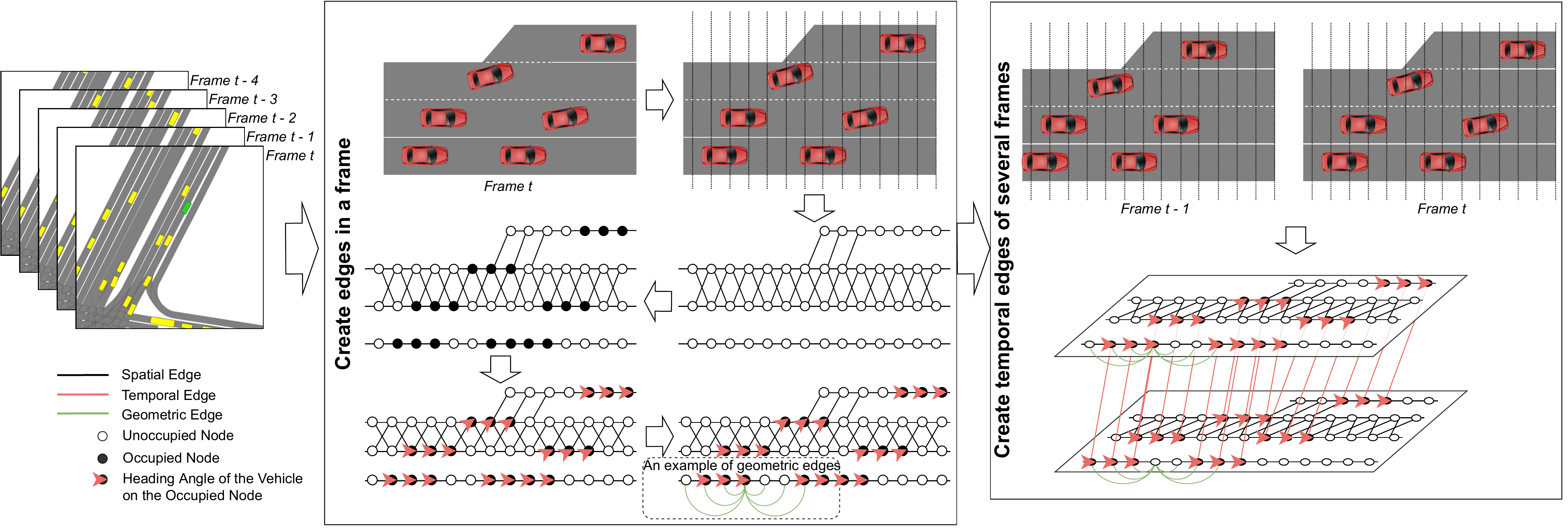}
    \vspace{-0.3cm}
    \caption{TOFG construction process. (1) Extract lane centerlines from raw map data and cut the lane centerlines into fine-grained lane segments. (2) Build lane graph based on extracted lane segments and their connections. (3) Build OFG for frame $t$. Compute occupancy flows given vehicles' trajectories and construct vehicle interaction edges and multi-scale geometric edges. (4) Connect OFGs in consecutive frames with temporal edges.}
    \label{fig:TOFG}
    \vspace{-0.6cm}
\end{figure*}

\subsection{Occupancy Flow Graph (OFG)}

Given the lane information provided by HD maps, an OFG is constructed based on the lane graph structure. Current HD maps are built based on the geometry of lanes and lane connectors, i.e., ramps or crossroads. \nop{Usually, each lane or lane connector is represented by a polyline which is an approximate geometry of its center line.} We first extract the geometry of the center line for each lane and connections among lanes, which are usually represented by polylines. Each center line can be represented as a sequence of lane segments, denoted as $L=(l_1, l_2, ... l_{N_{\rm seg}})$ with $N_{\rm seg}$ being the number of segments. Treating each lane segment as a graph node, all these lane segments form a lane graph. The lane segments are short \nop{since lanes are }for defining the detailed and fine curves or polygons of lanes\nop{ in an HD map}. These short lane segments serving as nodes allow the OFG to capture a more fine-grained map structure. To trade-off between the precision and complexity of the OFG, we set the average length of these lane segments as $0.3$ meters.

\nop{Generating occupancy graph is similar to generating occupancy grid. Instead of evenly spaced grid cells that divide the surrounding environment,} 
We use the above lane segments and lane graph to represent the environment and model the obstacles in it.
\textcolor{black}{We expand each lane segment $l=[(x_1, y_1)\rightarrow(x_2, y_2)]$ to a rectangle $R_l$ according to the lane width\nop{ $w_l$}. A lane segment\nop{ $l$} is occupied if its expanded rectangle $R_l$ intersects with the bounding box of a vehicle. If more than one vehicle's bounding box intersects with $R_l$, then the one closest to or containing the centroid $(\frac{x_1+x_2}{2}, \frac{y_1+y_2}{2})$ of $R_l$ is considered to be the occupant.}

Inspired by~\cite{mahjourian2022occupancy} which proposes OFF to extract the motion and future movement of occupant vehicles,
we further extend OFF to graphs to effectively extract interactions and integrate more informative features. With the lane graph structure described above, we could build an OFG as $G_{\rm OF} = \{\mathcal{V}, E\}$, where \textcolor{black}{$\mathcal{V} = \{u_1, u_2, ..., u_n\}$} is the set of $n$ nodes, and $E$ is the set of edges. For clearer distinction, we consider the following features to construct the node in the graph.\\
\textbf{Lane segment features:} Each lane segment $l$ starting from $(x_1, y_1)$ to $(x_2, y_2)$, is represented by its midpoint coordinates $(\frac{x_1+x_2}{2}, \frac{y_1+y_2}{2})$ and its vector form $(x_2-x_1, y_2-y_1)$.\\
\textbf{Occupant vehicle features:} Each vehicle that occupies lane segments includes an occupancy value $O \in \{0, 1\}$ and an occupancy flow vector \textcolor{black}{$(-v_x, -v_y, \theta, \omega)$, where $v = (v_x, v_y)$ is the velocity of this vehicle}, $\theta$ and $\omega$ are the yaw and yaw rate of the occupant vehicle, respectively. Here we follow the backward flow representation in~\cite{mahjourian2022occupancy} and use negative velocity. \\
The above features are used to form the node of the graph.
The edges that connect the nodes are defined as follows.\\
\textbf{Geometric edges:} Two adjacent lane segments are connected using an edge following the geometric edges of the lane graph extracted from the HD map.\\
\textbf{Multi-scale geometric edges:} In order to capture long range dependencies, we extend normal geometric edges and propose the multi-scale geometric edges. They are a union of all $n$-scale geometric edges, where $n=\{1, 2, ..., N_{\rm scale}\}$, and an $n$-scale geometric edge connects the current node with the node $n$ hops away. In this paper, we choose $N_{\rm scale}=4$.\\
\textbf{Vehicle interaction edges:} \nop{In real world driving scenarios, agents such as vehicles and pedestrians usually interact with each other. Therefore, }We design vehicle interaction edges to integrate the vehicle-to-vehicle interactions\nop{consider such interactions} in our OFG. We assume that there would be interactions between every two vehicles when the distance between them is less than a threshold, which is empirically set to 100 meters in this paper. For two interacting vehicles, the vehicle interaction edges connect the nodes between the bounding boxes of the two vehicles. Specifically, if vehicle A occupies a set of nodes $\mathcal{V}_A = \{u_1, u_2, \cdots , u_m\}$ and vehicle B occupies $\mathcal{V}_B=\{u_1', u_2', \cdots , u_n'\}$, where $m \le n$, the set of the vehicle interaction edges $E_{AB}$ are the injection from $\mathcal{V}_A$ to $\mathcal{V}_B$ without loss of generality, namely $E_{AB}=\{(u_1, u_1'), (u_2, u_2'), ..., (u_m, u_m')\}$.

The creation of an OFG given a frame of the data is illustrated in the first part of Figure~\ref{fig:TOFG}.

\subsection{Temporal Occupancy Flow Graph (TOFG)}

\begin{figure*}[t]
    \centering
    \includegraphics[width=0.95\textwidth]{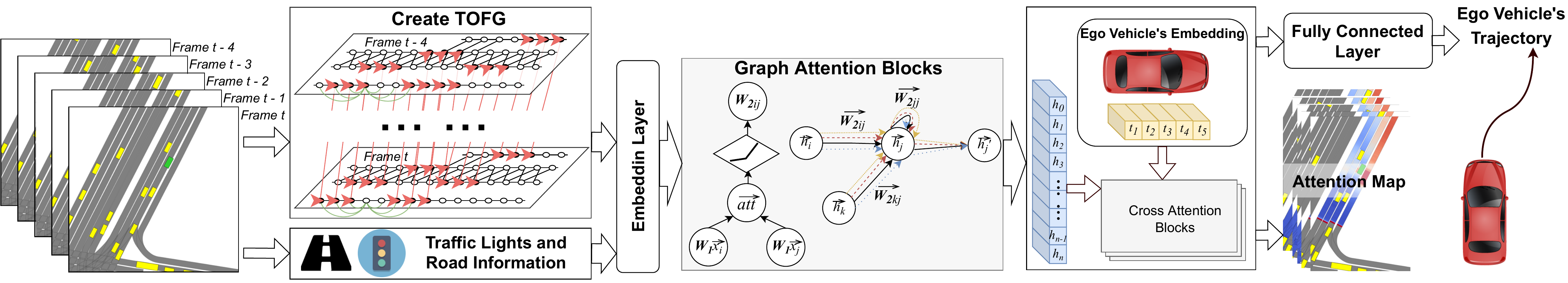}
    \vspace{-0.2cm}
    \caption{Architecture of TOFG-GAT.}
    \label{fig:gattofg}
    \vspace{-0.5cm}
\end{figure*}

Existing literature has proved that the historical trajectories of a moving object contribute to its future action prediction~\cite{zablocki2021explainability}. 
Moreover, human drivers, even autonomous vehicles, have reaction time due to the time lag of perception and information processing so that a driving behavior may be resulted from some events seconds ago.
Therefore, a single OFG cannot effectively handle such hysteretic time dependencies between perceived environment and observed driving decisions. To this end, we further propose Temporal Occupancy Flow Graph (TOFG) $G_{\rm TOF}$, the combination of multiple OFGs $G_{\rm OF}^{(t)}$ in a given time horizon. 

The \textbf{temporal edges} are introduced to connect the nodes between two OFGs in consecutive frames so that the temporal information could be propagated. For a vehicle which occupies $\mathcal{V}_t=\{u_1^{(t)}, u_2^{(t)}, ... , u_m^{(t)}\}$ at time $t$, if it exists in the previous OFG at time $t-1$ and occupies $\mathcal{V}_{t-1}=\{u_1^{(t-1)}, u_2^{(t-1)}, ... , u_n^{(t-1)}\}$, we conduct the following two operations to construct the temporal edges. First, we transform the coordinates of the occupied nodes to relative coordinates using the relative frame of the vehicle. Then, temporal edges are built to connect all occupied nodes at time $t$ to their closest nodes at time $t-1$ in the relative frame \textcolor{black}{in a similar way to vehicle interaction edges}. The creation of TOFG from several consecutive OFGs is shown in the right part of Figure~\ref{fig:TOFG}.

\section{GAT Model with TOFG} \label{sec:model}

Our proposed TOFG can be applied in many tasks, among which two representatives are motion planning and trajectory prediction. 
By unifying the map information and vehicles' trajectories into a homogeneous graph, TOFG enables the downstream models to encode consistent interaction\nop{between the ego vehicle and its surrounding environment} among vehicles and lanes using a simplified model structure without sacrificing accuracy. 
To facilitate the subsequent demonstration of these merits, we propose
\textbf{TOFG-GAT} for the downstream tasks and introduce the detailed model design in the rest of this section.




\subsection{Model Design}

    
    
    
    
    


Considering that the attention mechanism performs a great capability of capturing the relations among nodes, we apply a GAT-based model with our proposed TOFG, i.e., \textbf{TOFG-GAT} to predict the trajectory of the ego vehicle. As shown in Figure~\ref{fig:gattofg}, our model consists of three components. The first component is TOFG construction based on trajectory and HD map inputs. \textcolor{black}{The second component is a GAT network. \nop{Besides the created TOFG, the road information is also required to input to the GAT. The road information } To enrich semantic information, we embed some other road information with TOFG, which includes the following features: (1) the traffic light status, which could be one of \{red, yellow, green, none\}, and (2) the on-route status of lane segments, which could be ``on route'' or ``off route''.\nop{ Specifically, if a lane segment in a road block is on the ground truth trajectory, then it is ``on route". Otherwise, it is ``off route".}} Given the above inputs, the GAT network is responsible for extracting spatial and temporal interaction features from the environment. Note that the graph attention layers adopted here are slightly different from the one in~\cite{velickovic2017gat}. Given a node $i$ in TOFG, the features from its adjacent nodes $j$ are aggregated as Equation~\ref{eq:gat}.
\begin{equation} \label{eq:gat}
    h'_i = h_i + \sum_j \phi((h_i||h_j) W_1) W_2
\end{equation}
where $h_i$ is the embedding of node $u_i$, $W_1$, $W_2$ are trainable weight matrices, $\phi$ is the combination of layer normalization and ReLU, symbol $||$ is concatenation.

\nop{\textcolor{red}{Regarding the method, in III.A, the author claimed that "to make the derived attention scores informative, we only use one multi-head attention layer." I do not understand why one MHA layer makes the derived attention scores informative. Are you saying that having more layers causes an optimization challenge or computation overhead?}

\textcolor{black}{Multiple MHAs could work, but each MHA has its own attention map. I think if we use multiple MHAs, then each MHA's attention map may focus on different aspects, thus less comprehensive than a single MHA's attention map.}}

The third component is a cross-attention model to calculate the ego vehicle's attention score to every node of TOFGs in order to plan its future trajectory. \nop{To make the derived attention scores informative, }We only use one multi-head attention layer~\cite{vaswani2017attention} with $N_{\rm head}=4$ \textcolor{black}{so that the attention scores can be summarized in one attention map, which can provide comprehensive information}. The computing process of the multi-head cross-attention layer is defined in Equation~\ref{eq:cross_att}.
\begin{equation} \label{eq:cross_att}
\begin{small}
\begin{aligned}
    &{\rm Attention}(Q, K, V) = {\rm softmax}(\frac{QK^T}{\sqrt{d_k}})V \\
    &{\rm head}_i = {\rm Attention}(Q(h_{\rm ego}), K(h_{\rm tofg}), V(h_{\rm tofg})) \\
    &y_{\rm att} = (\Vert_{i=1}^{N_{\rm head}} {\rm head}_i) W^{att}
\end{aligned}
\end{small}
\end{equation}
where $h_{\rm ego}$ and $h_{\rm tofg}$ are the embedding of the ego vehicle and the embedding matrix of the TOFG nodes, respectively; $Q$, $K$, and $V$ are query, key, and value networks, respectively, and $d_k$ is the output dimension of $Q$ and $K$. and $W^{att}$ is a trainable weight matrix. The output of the cross-attention module $y_{\rm att}$ is fed into a Muti-Layer Perceptron (MLP) layer to generate \nop{the final output, which is }a discrete trajectory \nop{$\overline{\pi} =\{(\overline{x}_1, \overline{y}_1), (\overli\nop\nop{x}_2, \overline{y}_2), \cdots, (\overline{x}_T, \overline{y}_T)\}$ }for several future\nop{ $T$} time steps.

\subsection{Training Loss}

We follow the classic imitation learning \nop{(behavior cloning) }setting \cite{hussein2017imitation} and use imitation loss as the loss function for our model. Denote $\overline{\pi} =\{(\overline{x}_1, \overline{y}_1), (\overline{x}_2, \overline{y}_2), \cdots, (\overline{x}_T, \overline{y}_T)\}$ as the trajectory planned by our model and $\pi =\{(x_1, y_1), (x_2, y_2), \cdots, (x_T, y_T)\}$ as the ground truth trajectory of the vehicle in the dataset. The imitation loss is defined as $L_{\rm imitation} = \sum_{t=1}^T \sqrt{(\overline{x}_t - x_t)^2 + (\overline{y}_t - y_t)^2}$.

\section{Experiment} \label{sec:experiment}

To demonstrate the superiority of TOFG-GAT in the tasks of trajectory prediction and motion planning, we compare it with the following baseline models: LaneGCN~\cite{liang2020lanegcn}, mmTransformer~\cite{liu2021multimodal}, and HiVT~\cite{zhou2022hivt}. \textcolor{black}{Note that we slightly change the output layer of the original HiVT to fit in our experiment settings.} All the three baseline models are constructed based on GNNs while different attention mechanisms are adopted to capture the interactions among vehicles and lanes. 

To train and evaluate both the proposed model and the baselines, NuPlan~\cite{nuplan} mini dataset is utilized. NuPlan provides 1500 hours of human driving data from 4 cities across the US and Asia with diverse traffic patterns, and the mini dataset is a teaser version of the full dataset and contains about 2.5\% of all data (around 40 hours).
\nop{To promote reproducibility, more details of implementing our proposed model can be found below. The embedding module from TOFG to the downstream GAT is a four-layer MLP. Our GAT contains four graph attention layers described in Section~\ref{sec:model}. The cross-attention module contains only one cross-attention layer with its query, key, and value network being a single fully-connected layer. Finally, a three-layer MLP to output a trajectory at the end of our model. The size of all hidden layers in our model is 128. }
We choose trajectory prediction and motion planning tasks to illustrate the performance of our TOFG-GAT. We randomly extract 5000 scenario samples from NuPlan mini dataset and split them into training (70\%), validation (15\%), and testing set (15\%). Each of the models is trained on the training set for $60$ epochs. We use Adam optimizer with a learning rate of $1\times10^{-5}$ and set batch size as $3$ due to the limitation of GPU memory. We select the checkpoint that performs the best in validation from training epochs for testing.

\subsection{Trajectory Prediction} \label{sec:traj_pred}

We perform a single-agent trajectory prediction task to illustrate our motivation example numerically and the superiority of TOFG-GAT. 
The problem of the single-agent trajectory prediction is to predict a $6$-second future trajectory of ego vehicle $\hat{\pi}_0 = (\hat{s}_0^{t+1}, \hat{s}_0^{t+2}, ..., \hat{s}_0^{t+H})$, given the $1.5$-second historical trajectory of ego and surrounding vehicles $\Pi=(\pi_0, \pi_1, ..., \pi_m)$ and map information $\mathcal{M}$, where $\pi_i=(s_i^{t-T}, s_i^{t-T+1}, ..., s_i^{t})$ is the historical trajectory of vehicle $i$, $s_i^{t}=(x_i^t, y_i^t, \theta_i^t)$ is the state of vehicle $i$ at time $t$, and $H$ and $T$ are the numbers of time steps in prediction horizon and past time horizon, respectively. Here, we set $H=12$ and $T=5$ following NuPlan's setting.
We use the following four metrics to evaluate the performance: average displacement error (ADE), average heading error (AHE), final displacement error (FDE), and final heading error (FHE).
They are formally defined as Equation~\ref{eq:ade_fde_displacement}.
\begin{equation} \label{eq:ade_fde_displacement}
\vspace{-0.1cm}
\begin{small}
\begin{aligned}
    {\rm ADE} &= \frac{1}{H} \sum_{i=0}^{H} \Vert \hat{s}_0^{t+i} - s_0^{t+i} \Vert_2,\ 
    {\rm FDE} = \Vert \hat{s}_0^{t+H} - s_0^{t+H} \Vert_2 \\
    {\rm AHE} &= \frac{1}{H} \sum_{i=0}^{H} \left| \hat{\theta}_0^{t+i} - \theta_0^{t+i} \right|, \ 
    {\rm FHE} = \left| \hat{\theta}_0^{t+H} - \theta_0^{t+H} \right|
\end{aligned}
\end{small}
\vspace{-0.1cm}
\end{equation}

First, we verify our motivation by training the aforementioned two LaneGCN models in Section \ref{sec:motivation_example}: the vanilla LaneGCN (attention layer order:\textless A2L, L2A, A2A\textgreater) and the modified LaneGCN (attention layer order: \textless A2A, A2L, L2A\textgreater). Since the lane change is a more complicated and challenging scenario, different from the conventional training procedures, we extract 5000 lane-changing scenarios from NuPlan mini dataset for a better illustration of the impact of attention order. The extracted scenarios are divided into training (70\%), validation (15\%), and testing set (15\%). Both models are trained on the training set for $30$ epochs and tested on the testing set. The result of the verification experiment is shown in Table \ref{tab:vrf_exp}. We can see that except for AHE, the overall performance of the modified LaneGCN is better than the original one, verifying our motivation that there is no one-size-fits-all attention order.

Next, we compare the performance of TOFG-GAT and the baselines on the trajectory prediction task. The training setting of the prediction task follows the aforementioned conventional one in the beginning of this section. The result of the trajectory prediction task using our model and baselines are shown in Table~\ref{tab:prediction}. It can be seen from the table that TOFG-GAT performs the best in ADE, AHE, and FHE. And our FDE is preceded only by mmTransformer. Specifically, our model has the smallest heading errors among baseline models, implying that our model performs better in terms of both position and heading orientation.

\begin{table} 
\caption{Trajectory prediction performance comparison achieved by two different LaneGCNs.}
\vspace{-0.2cm}
\label{tab:vrf_exp}
\centering
\fontsize{8}{9}\selectfont
\begin{tabular}{ccc}
    \toprule
    \textbf{Metric} & \textbf{Vanilla LaneGCN} & \textbf{Modified LaneGCN} \\
    \midrule
    \textbf{ADE (m)} & 2.348 & \textbf{2.235} \\
    \textbf{AHE (rad)} & \textbf{0.0851} & 0.0889 \\
    \textbf{FDE (m)} & 5.650 & \textbf{5.231} \\
    \textbf{FHE (rad)} & 0.1353 & \textbf{0.1324} \\
    \bottomrule
\end{tabular}
\end{table}

\begin{table} 
\caption{Trajectory prediction performance comparison achieved by TOFG-GAT and baselines.}
\vspace{-0.2cm}
\label{tab:prediction}
\fontsize{8}{9}\selectfont
\begin{tabular}{cp{0.9cm}p{0.9cm}p{0.9cm}p{0.9cm}}
    \toprule
    \textbf{Metric} & \textbf{TOFG-GAT} & \textbf{LaneGCN} & \textbf{mmTrans} & \textbf{HiVT} \\
    \midrule
    \textbf{ADE (m)} & \textbf{1.818} & 2.258 & 1.836 & 3.5571 \\
    \textbf{AHE (rad)} & \textbf{0.0463} & 0.0721 & 0.1133 & 0.1014 \\
    \textbf{FDE (m)} & 4.538 & 5.063 & \textbf{4.210} & 8.4658 \\
    \textbf{FHE (rad)} & \textbf{0.0818} & 0.1094 & 0.1421 & 0.1340 \\
    \textbf{Num parameters} & \textbf{542K} & 1.9M & 1.6M & 1.7M \\
    \bottomrule
\end{tabular}
\vspace{-0.6cm}
\end{table}

\subsection{Motion Planning}

To compare our model with the baselines in the motion planning task, we adopt close loop simulation, which provides a more in-depth evaluation since compounding errors during simulation largely affect future observations and could significantly diverge from the ground truth. 
In the close loop simulation, each model is required to plan a trajectory $\hat{\pi}^t$ every $k$ ms for controlling the ego vehicle given historical trajectory of ego and surrounding vehicles $\Pi$ and map information $\mathcal{M}$ at time $t$. 
We assume that the ego vehicle can perfectly follow $\hat{\pi}^t$ and move to the corresponding position $s^{t+k}$ on $\hat{\pi}$ at time step $t+k$. As for surrounding vehicles, we simply replay their driving behaviors from the dataset. For safety concerns, we add an extra auto-correction mechanism for ego vehicle to avoid collision or driving out of the road.

We randomly select $50$ simulation scenarios from NuPlan mini dataset. Each simulation lasts for $20$ seconds. For each scenario, there is a goal state $s_{\rm goal} = (x_{\rm goal}, y_{\rm goal}, \theta_{\rm goal})$, which is the true final state of ego vehicle.
Given the trajectory $\pi_{\rm v}=(\overline{s}^{0}, ..., \overline{s}^{T_s})$ of ego vehicle planned by the motion planning model and the ground truth one $\pi_{\rm expert}=(s^{0}, ..., s^{T_s})$, where $T_s$ is the duration time of the simulation. We use the following metrics to evaluate motion planning.\\
\textbf{(1) L2 distance to expert trajectory:} This metric evaluates the trajectory difference between expert trajectory and vehicle trajectory. We also include the heading error in this metric, with a weight factor $w_{\theta}=2.5$. The metric could be formulated as $M_{L2} = \sum_{t=0}^{T_s}(\Vert (\overline{x}^t - x^t, \overline{y}^{t} - y^{t}) \Vert_2 + w_{\theta} \left| \hat{\theta}^{t} - \theta^{t} \right|)$.\\
\textbf{(2) Distance to goal:} This metric is the distance from each planned waypoint to the simulation goal, and a smaller value is better. We take the maximum, minimum, and average of this metric for comparison, i.e., $M_{\rm dist2goal} = \Phi(\{\Vert (\overline{x}^t - x_{\rm goal}, \overline{y}^t - y_{\rm goal}) \Vert_2\}_t), t=0, 1, ..., T_s$, where $\Phi$ is $\min$, $\max$, or $\rm avg$. \\
\textbf{(3) Distance progress to goal:} This metric measures the progress of the ego vehicle toward the goal position. It is calculated using $M_{\rm prog2goal} = \Vert (\overline{x}^0 - x_{\rm goal}, \overline{y}^0 - y_{\rm goal}) \Vert_2 - \Vert (\overline{x}^{T_s} - x_{\rm goal}, \overline{y}^{T_s} - y_{\rm goal}) \Vert_2$.\\
\textbf{(4) Distance progress along expert route:} This metric accumulates ego vehicle's progress distance along the expert route, denoted as $M_{prog2exp}$, and a higher value is better. Ego vehicle is on the expert route if it is on the same roadblock of expert trajectory or the skipped road block connected to the next roadblock of expert trajectory. 

The result of motion planning is shown in Table~\ref{tab:planning}. TOFG-GAT has an overall best performance in terms of these four metrics. It is only slightly worse than HiVT in L2 distance. The reason behind this is that the planning accuracy of the first few frames of HiVT is slightly better than ours. The superiority of our model in the other three metrics demonstrates that TOFG-GAT has a better understanding of the environment and the intention of the ego vehicle since our representation can better capture inter-vehicle interactions and vehicle-lane interactions. 
Additionally, TOFG-GAT has the shortest inference time among the four models.


\nop{\textcolor{red}{For tables in the experiments, it can be better to describe whether each metric is higher-the-better or lower-the-better since not all metrics are prediction errors. In addition, it seems all experiments are conducted only once. If that is the case, can the author briefly discuss if that is due to the computation limit?}

\textcolor{black}{Yes, it is due to computation limit. But I do not understand why exp should be conducted multiple times.}}

\begin{table} 
\caption{Motion planning performance comparison achieved by TOFG-GAT and baselines.}
\vspace{-0.1cm}
\fontsize{7.5}{9}\selectfont
\begin{tabular}{p{1.7cm}p{0.8cm}p{0.9cm}p{0.9cm}p{0.9cm}p{0.9cm}}
\toprule
\multicolumn{2}{c}{\textbf{Metric}} & \textbf{TOFG-GAT} & \textbf{LaneGCN} & \textbf{mmTrans} & \textbf{HiVT} \\
\midrule
\multirow{4}*{\makecell{$M_{\rm L2}$\\(lower-the-better)}} & max & 50.7751 & 53.7209 & 60.6808 & \textbf{50.0980} \\
~ & mean & 23.7583 & 27.3519 & 28.7099 & \textbf{23.1862} \\
~ & max (yaw) & 52.1662 & 55.6454 & 62.9381 & \textbf{51.6376} \\
~ & mean (yaw) & 24.7023 & 28.7969 & 30.3691 & \textbf{24.1930} \\
\midrule
\multirow{3}*{\makecell{$M_{\rm dist2goal}$\\(lower-the-better)}} & max & \textbf{215.0541} & 217.4206 & 215.7068 & 215.2509 \\
~ & min & \textbf{118.7215} & 140.1732 & 190.6343 & 151.3336 \\
~ & mean & \textbf{166.5199} & 179.3244 & 202.9253 & 184.6144 \\
\midrule
\multirow{2}*{\makecell{$M_{\rm prog2goal}$\\(higher-the-better)}} & \nop{ego progress to goal value} absolute & \textbf{89.0107} & 66.3626 & 19.1848 & 59.1334 \\
~ & \nop{ego to expert relative progress to goal}relative & \textbf{0.8671} & 0.8312 & 0.5415 & 0.7435 \\
\midrule
\multirow{2}*{\makecell{$M_{\rm prog2exp}$\\(higher-the-better)}} 
 & \nop{ego} total \nop{progress along route value} & \textbf{81.0641} & 66.1283 & 26.1031 & 58.4340 \\
~ & \nop{ego expert progress along route} ratio & \textbf{0.7844} & 0.6887 & 0.3508 & 0.6121 \\
\midrule
\multirow{1}*{\makecell{Inference time (s)}} &  & \textbf{0.0122} & 0.0250 & 0.0125 & 0.0164 \\
\bottomrule
\end{tabular}
\label{tab:planning}
\vspace{-0.5cm}
\end{table}

\subsection{Attention Map Visualization and Analysis}

\begin{figure}
    \centering
    \subfigure[Rendered scenario]{
    \label{fig:scenario}
    \includegraphics[width=0.45\linewidth]{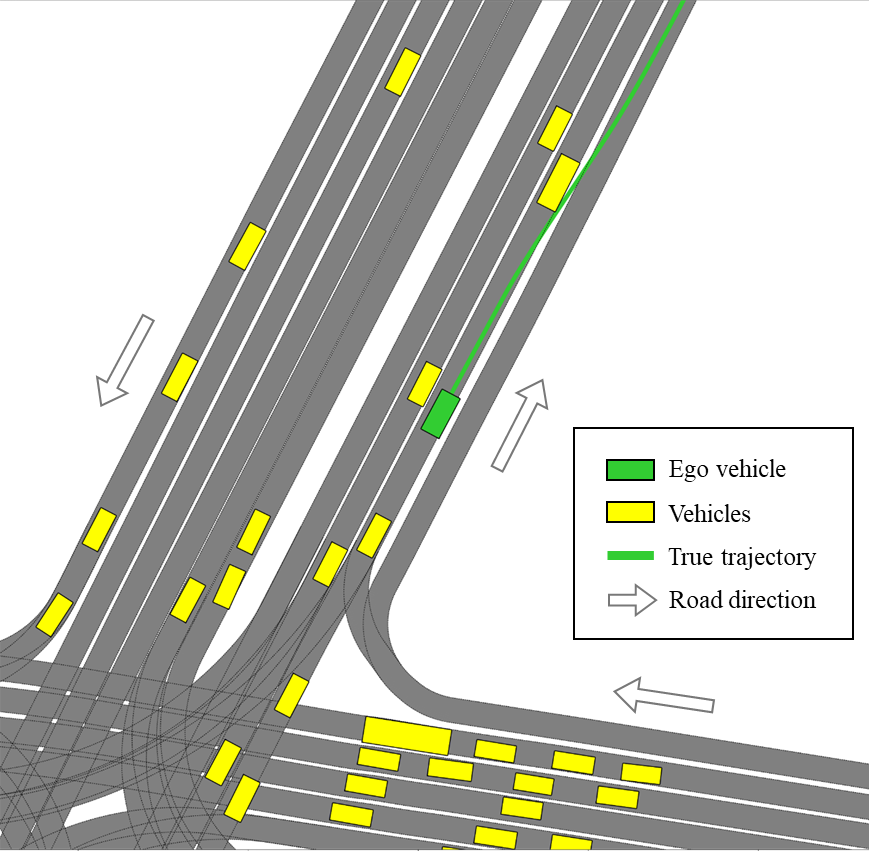}}
    \subfigure[Attention map by TOFG-GAT]{
    \label{fig:ours}
    \includegraphics[width=0.45\linewidth]{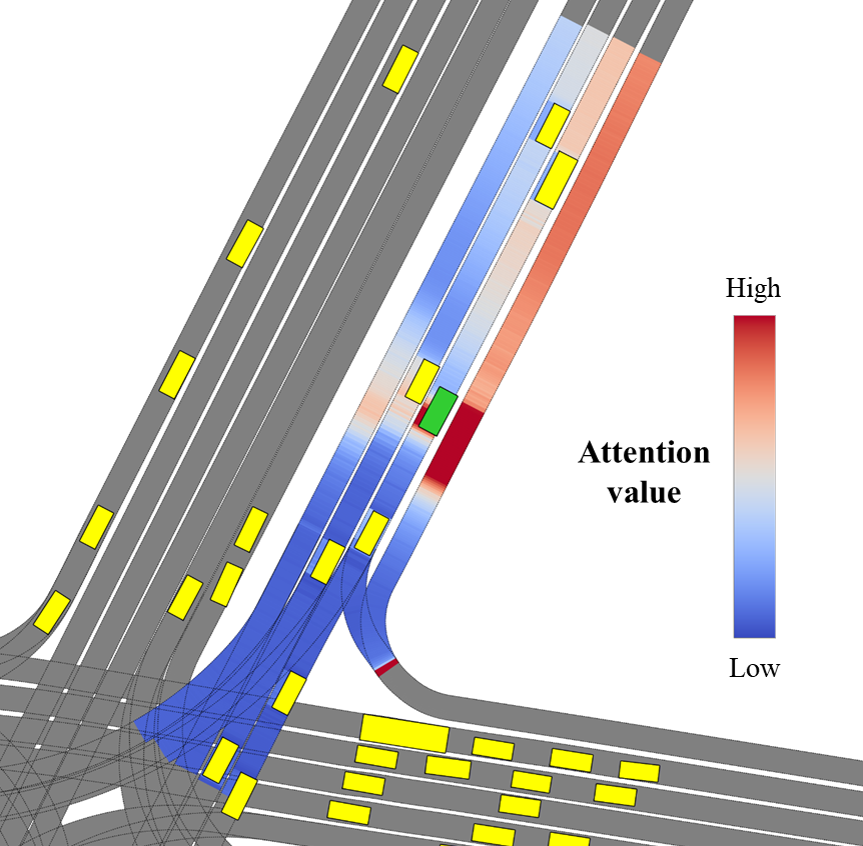}}
    \subfigure[A2L attention map by LaneGCN (ego perspective)]{
    \label{fig:lanegcn_a2l}
    \includegraphics[width=0.45\linewidth]{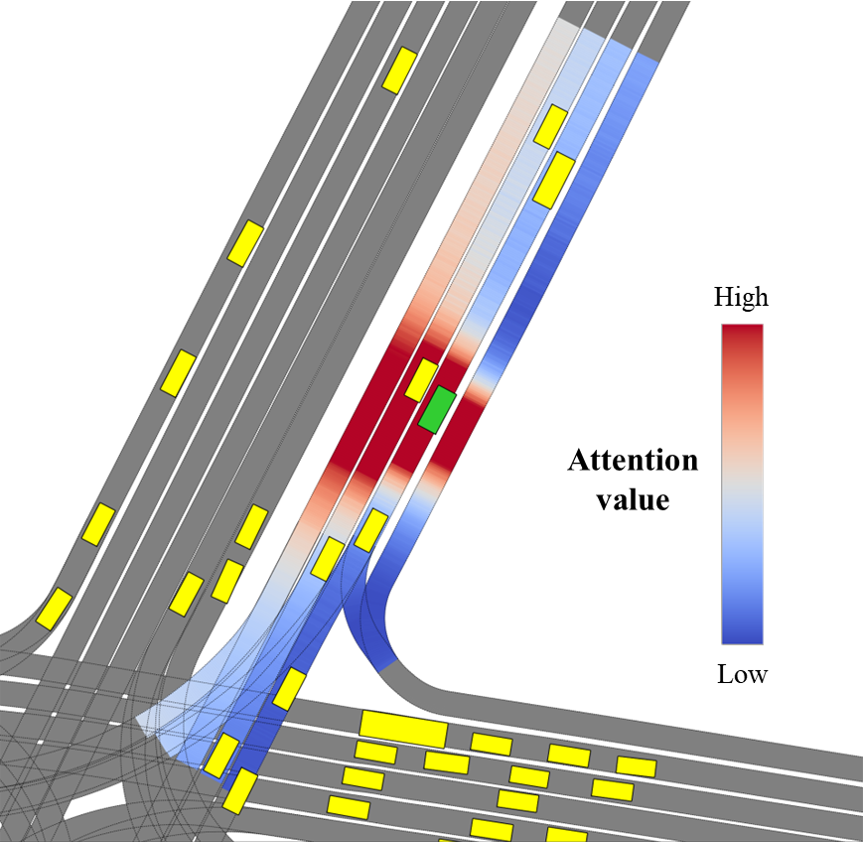}}
    \subfigure[A2A attention map by LaneGCN (ego perspective)]{
    \label{fig:lanegcn_a2a}
    \includegraphics[width=0.45\linewidth]{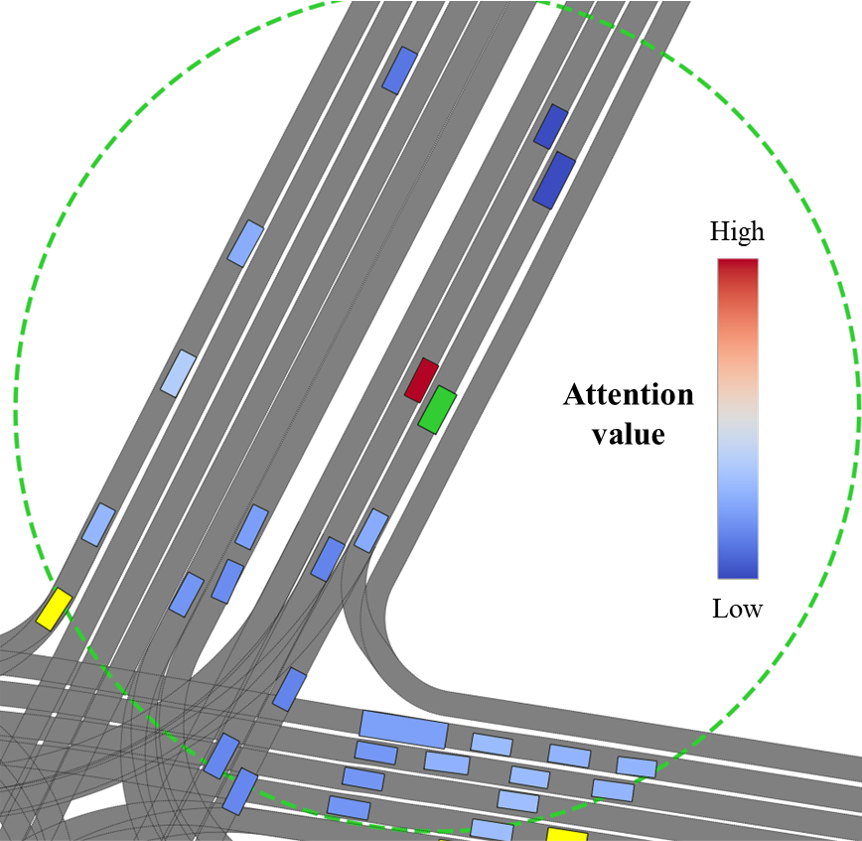}}
    \vspace{-0.1cm}
    \caption{Visualization of attention maps from TOFG-GAT and LaneGCN. The attentions of TOFG-GAT points out the target lane and places near the vehicles in interactions, while the attentions of LaneGCN are basically related to distance between ego and the attention subject.}
\label{fig:att_map}
\vspace{-0.6cm}
\end{figure}

Finally, we analyze the attention map extracted from TOFG-GAT and LaneGCN. Figure~\ref{fig:scenario} shows a lane-changing scenario extracted from NuPlan dataset: ego vehicle marked with green is changing to the lane on its right. The last frame's attention map of the cross-attention layer from TOFG-GAT and the ego vehicle's attention maps from A2L and A2A module of LaneGCN are shown using heat maps in Figure~\ref{fig:ours}, \ref{fig:lanegcn_a2l}, and \ref{fig:lanegcn_a2a}, respectively.


In Figure~\ref{fig:ours}, the attention values on the target lane are higher than those on other lanes. Also, the attention values on the lane segments around neighbor vehicles, which are likely to have interactions with the ego vehicle, are higher than others vehicles. Such pattern allows our model to make a similar choice to the ground truth while keeping ego vehicle from colliding with others.
The attention maps from LaneGCN are less informative than ours, as shown in Figure~\ref{fig:lanegcn_a2l} and \ref{fig:lanegcn_a2a}. The A2L attention is basically a monotonic increase function to distance between ego vehicle and the lane segment. Similarly, in A2A attention map, the attention value seems to be higher when the distance between the ego vehicle and the other vehicle become larger. Moreover, the vehicles interacting with the ego vehicle do not get larger attention value.
The above case studies well demonstrates that our model can obtain more informative attention than baseline models. 

\section{Conclusion} \label{sec:conclusion}
In this paper, we proposed a unified environment representation TOFG to include both the map information and trajectory of vehicles in a homogeneous graph. The lane segments are partitioned with a smaller length to construct a graph with fine-grained information. Such a unified and fine-grained graph can contribute to simultaneously and accurately capturing the interactions so that the performance of downstream models can be further improved. As the complex attention layers are no longer required to tackle different interactions of heterogeneous data, a simplified model TOFG-GAT with fewer parameters can achieve a competitive and even better performance than that of SOTA models\nop{ with less training time}. In future work, we plan to analyze the attention map numerically to summarize a common interaction-aware guideline that can be applied to all types of models of autonomous driving tasks.

\clearpage
\bibliographystyle{IEEEtran}
\bibliography{IEEEexample}

\begin{thebibliography}{10}
\providecommand{\url}[1]{#1}
\csname url@rmstyle\endcsname
\providecommand{\newblock}{\relax}
\providecommand{\bibinfo}[2]{#2}
\providecommand\BIBentrySTDinterwordspacing{\spaceskip=0pt\relax}
\providecommand\BIBentryALTinterwordstretchfactor{4}
\providecommand\BIBentryALTinterwordspacing{\spaceskip=\fontdimen2\font plus
\BIBentryALTinterwordstretchfactor\fontdimen3\font minus
  \fontdimen4\font\relax}
\providecommand\BIBforeignlanguage[2]{{%
\expandafter\ifx\csname l@#1\endcsname\relax
\typeout{** WARNING: IEEEtran.bst: No hyphenation pattern has been}%
\typeout{** loaded for the language `#1'. Using the pattern for}%
\typeout{** the default language instead.}%
\else
\language=\csname l@#1\endcsname
\fi
#2}}

\bibitem{yurtsever2020survey}
E.~Yurtsever, J.~Lambert, A.~Carballo, and K.~Takeda, ``A survey of autonomous
  driving: Common practices and emerging technologies,'' \emph{IEEE access},
  vol.~8, pp. 58\,443--58\,469, 2020.

\bibitem{grigorescu2020survey}
S.~Grigorescu, B.~Trasnea, T.~Cocias, and G.~Macesanu, ``A survey of deep
  learning techniques for autonomous driving,'' \emph{Journal of Field
  Robotics}, vol.~37, no.~3, pp. 362--386, 2020.

\bibitem{ye2022gsan}
L.~Ye, Z.~Wang, X.~Chen, J.~Wang, K.~Wu, and K.~Lu, ``Gsan: Graph
  self-attention network for learning spatial–temporal interaction
  representation in autonomous driving,'' \emph{IEEE Internet of Things
  Journal}, vol.~9, no.~12, pp. 9190--9204, 2022.

\bibitem{interaction1}
G.~Markkula, R.~Madigan, D.~Nathanael, E.~Portouli, Y.~M. Lee, A.~Dietrich,
  J.~Billington, A.~Schieben, and N.~Merat, ``Defining interactions: a
  conceptual framework for understanding interactive behaviour in human and
  automated road traffic,'' \emph{Theoretical Issues in Ergonomics Science},
  vol.~21, no.~6, pp. 728--752, 2020.

\bibitem{interaction2}
N.~Deo and M.~M. Trivedi, ``Convolutional social pooling for vehicle trajectory
  prediction,'' in \emph{2018 IEEE/CVF Conference on Computer Vision and
  Pattern Recognition Workshops (CVPRW)}, 2018, pp. 1549--15\,498.

\bibitem{velickovic2017gat}
P.~Velickovic, G.~Cucurull, A.~Casanova, A.~Romero, P.~Lio, and Y.~Bengio,
  ``Graph attention networks,'' \emph{stat}, vol. 1050, p.~20, 2017.

\bibitem{zhou2022hivt}
Z.~Zhou, L.~Ye, J.~Wang, K.~Wu, and K.~Lu, ``Hivt: Hierarchical vector
  transformer for multi-agent motion prediction,'' in \emph{Proceedings of the
  IEEE/CVF Conference on Computer Vision and Pattern Recognition (CVPR)}, 2022.

\bibitem{liu2021multimodal}
Y.~Liu, J.~Zhang, L.~Fang, Q.~Jiang, and B.~Zhou, ``Multimodal motion
  prediction with stacked transformers,'' \emph{Computer Vision and Pattern
  Recognition}, 2021.

\bibitem{Gu_2021_ICCV}
J.~Gu, C.~Sun, and H.~Zhao, ``Densetnt: End-to-end trajectory prediction from
  dense goal sets,'' in \emph{Proceedings of the IEEE/CVF International
  Conference on Computer Vision (ICCV)}, October 2021, pp. 15\,303--15\,312.

\bibitem{ngiam2021scene}
J.~Ngiam, B.~Caine, V.~Vasudevan, Z.~Zhang, H.-T.~L. Chiang, J.~Ling,
  R.~Roelofs, A.~Bewley, C.~Liu, A.~Venugopal, \emph{et~al.}, ``Scene
  transformer: A unified architecture for predicting multiple agent
  trajectories,'' \emph{arXiv preprint arXiv:2106.08417}, 2021.

\bibitem{NEURIPS2021_224e5e49}
C.~Yu and S.~Gao, ``Reducing collision checking for sampling-based motion
  planning using graph neural networks,'' in \emph{Advances in Neural
  Information Processing Systems}, M.~Ranzato, A.~Beygelzimer, Y.~Dauphin,
  P.~Liang, and J.~W. Vaughan, Eds., vol.~34.\hskip 1em plus 0.5em minus
  0.4em\relax Curran Associates, Inc., 2021, pp. 4274--4289.

\bibitem{interaction-planning}
B.~Ichter, E.~Schmerling, T.-W.~E. Lee, and A.~Faust, ``Learned critical
  probabilistic roadmaps for robotic motion planning,'' 2019.

\bibitem{liang2020lanegcn}
M.~Liang, B.~Yang, R.~Hu, Y.~Chen, R.~Liao, S.~Feng, and R.~Urtasun, ``Learning
  lane graph representations for motion forecasting,'' in \emph{European
  Conference on Computer Vision}.\hskip 1em plus 0.5em minus 0.4em\relax
  Springer, 2020, pp. 541--556.

\bibitem{vaswani2017attention}
A.~Vaswani, N.~Shazeer, N.~Parmar, J.~Uszkoreit, L.~Jones, A.~N. Gomez,
  {\L}.~Kaiser, and I.~Polosukhin, ``Attention is all you need,''
  \emph{Advances in neural information processing systems}, vol.~30, 2017.

\bibitem{learn-planning1}
A.~Kuefler, J.~Morton, T.~Wheeler, and M.~Kochenderfer, ``Imitating driver
  behavior with generative adversarial networks,'' in \emph{2017 IEEE
  Intelligent Vehicles Symposium (IV)}.\hskip 1em plus 0.5em minus 0.4em\relax
  IEEE Press, 2017, p. 204–211.

\bibitem{v2v2}
N.~Deo and M.~M. Trivedi, ``Multi-modal trajectory prediction of surrounding
  vehicles with maneuver based lstms,'' in \emph{2018 IEEE Intelligent Vehicles
  Symposium (IV)}.\hskip 1em plus 0.5em minus 0.4em\relax IEEE Press, 2018, p.
  1179–1184.

\bibitem{v2v3}
X.~Li, X.~Ying, and M.~C. Chuah, ``Grip: Graph-based interaction-aware
  trajectory prediction,'' in \emph{2019 IEEE Intelligent Transportation
  Systems Conference (ITSC)}, 2019, pp. 3960--3966.

\bibitem{gohome}
T.~Gilles, S.~Sabatini, D.~Tsishkou, B.~Stanciulescu, and F.~Moutarde,
  ``Gohome: Graph-oriented heatmap output for future motion estimation,'' in
  \emph{2022 International Conference on Robotics and Automation (ICRA)}, 2022,
  pp. 9107--9114.

\bibitem{v2l1}
J.~Mercat, T.~Gilles, N.~E. Zoghby, G.~Sandou, D.~Beauvois, and G.~P. Gil,
  ``Multi-head attention for multi-modal joint vehicle motion forecasting,''
  2019.

\bibitem{Gao_2020_CVPR}
J.~Gao, C.~Sun, H.~Zhao, Y.~Shen, D.~Anguelov, C.~Li, and C.~Schmid,
  ``Vectornet: Encoding hd maps and agent dynamics from vectorized
  representation,'' in \emph{Proceedings of the IEEE/CVF Conference on Computer
  Vision and Pattern Recognition (CVPR)}, June 2020.

\bibitem{Chang_2019_CVPR}
M.-F. Chang, J.~Lambert, P.~Sangkloy, J.~Singh, S.~Bak, A.~Hartnett, D.~Wang,
  P.~Carr, S.~Lucey, D.~Ramanan, and J.~Hays, ``Argoverse: 3d tracking and
  forecasting with rich maps,'' in \emph{Proceedings of the IEEE/CVF Conference
  on Computer Vision and Pattern Recognition (CVPR)}, June 2019.

\bibitem{interaction3}
W.~Schwarting, A.~Pierson, J.~Alonso-Mora, S.~Karaman, and D.~Rus, ``Social
  behavior for autonomous vehicles,'' \emph{Proceedings of the National Academy
  of Sciences}, vol. 116, no.~50, pp. 24\,972--24\,978, 2019.

\bibitem{mahjourian2022occupancy}
R.~Mahjourian, J.~Kim, Y.~Chai, M.~Tan, B.~Sapp, and D.~Anguelov, ``Occupancy
  flow fields for motion forecasting in autonomous driving,'' \emph{IEEE
  Robotics and Automation Letters}, vol.~7, no.~2, pp. 5639--5646, 2022.

\bibitem{zablocki2021explainability}
{\'E}.~Zablocki, H.~Ben-Younes, P.~P{\'e}rez, and M.~Cord, ``Explainability of
  vision-based autonomous driving systems: Review and challenges,'' \emph{arXiv
  preprint arXiv:2101.05307}, 2021.

\bibitem{hussein2017imitation}
A.~Hussein, M.~M. Gaber, E.~Elyan, and C.~Jayne, ``Imitation learning: A survey
  of learning methods,'' \emph{ACM Computing Surveys (CSUR)}, vol.~50, no.~2,
  pp. 1--35, 2017.

\bibitem{nuplan}
K.~T. e.~a. H.~Caesar, J.~Kabzan, ``Nuplan: A closed-loop ml-based planning
  benchmark for autonomous vehicles,'' in \emph{CVPR ADP3 workshop}, 2021.

\end{thebibliography}

\end{document}